\documentclass[11pt]{article}

\usepackage{microtype} 
\usepackage{booktabs}  
\usepackage{url}  
\usepackage{csquotes}

\usepackage{amsmath}
\usepackage{amsthm}

\usepackage{graphicx} 

\usepackage{float}

\usepackage[final]{automl}

\usepackage{natbib}
\title{SNAC-Pack 2.0: Scaled-Out Surrogate\\Neural Architecture Codesign}

\author[1]{\nameemail{Jason Weitz}{jdweitz@ucsd.edu}}
\author[2,1]{\nameemail{Dmitri Demler}{ddemler@ethz.ch}}
\author[3]{\nameemail{Benjamin Hawks}{bhawks@fnal.gov}}
\author[4]{\nameemail{Aaron Wang}{aaronw5@uic.edu}}
\author[3]{\nameemail{Nhan Tran}{ntran@fnal.gov}}
\author[1]{\nameemail{Javier Duarte}{jduarte@ucsd.edu}}

\affil[1]{University of California San Diego}
\affil[2]{ETH Zurich}
\affil[3]{Fermi National Accelerator Laboratory }
\affil[4]{University of Illinois Chicago}

\hypersetup{%
  pdfauthor={Jason Weitz, Dmitri Demler, Benjamin Hawks, Aaron Wang, Nhan Tran, Javier Duarte}, 
  pdftitle={SNAC-Pack 2.0: Scaled-Out Surrogate Neural Architecture Codesign},
  pdfsubject={AutoML Conference Submission},
  pdfkeywords={neural architecture search, FPGA, hardware-aware}
}

\begin{document}

\maketitle

\begin{abstract}
Neural architecture search (NAS) is a powerful approach for automating model design, but existing methods often optimize for accuracy alone or rely on proxy metrics such as bit operations (BOPs) that correlate poorly with hardware cost.
This gap is particularly large for FPGA deployment, where cost is dominated by a multi-dimensional budget of lookup tables, DSPs, flip-flops, BRAM, and latency.
We present the Surrogate Neural Architecture Codesign Package (SNAC-Pack), an open-source AutoML framework for hardware-aware neural architecture codesign and end-to-end FPGA deployment.
SNAC-Pack runs a multi-objective global search with Optuna and NSGA-II, loading trials to a shared SQLite store that enables parallel workers across compute nodes.
A hardware surrogate model outputs per-trial resource and latency estimates, avoiding the synthesis cost that would otherwise dominate the search loop.
A local search stage then applies quantization-aware training (QAT) together with iterative magnitude pruning in a combined compression loop, after which the final model is synthesized to FPGA firmware via the hls4ml Python library.
A YAML configuration and an optional agentic frontend let users run the pipeline on new datasets without modifying the framework.
We demonstrate SNAC-Pack on jet classification at the Large Hadron Collider and superconducting qubit readout, discovering compact Pareto-optimal architectures that either match baseline task performance while reducing FPGA resource utilization or substantially reduce hardware utilization with only modest decreases in task performance. In the qubit readout case, SNAC-Pack also reduces the design space exploration process from months of manual fine-tuning to hours of automated search.
\end{abstract}

\section{Introduction}
\label{sec:intro}

Machine learning models now achieve state-of-the-art performance across many domains, but deploying them in resource-constrained environments such as edge devices, embedded accelerators, and low-latency triggers remains a substantial engineering challenge.
FPGAs are a natural target in such settings because they offer microsecond-scale inference and reconfigurable parallelism, but they impose a multi-dimensional resource budget spanning lookup tables (LUTs), digital signal processors (DSPs), block RAM (BRAM), flip-flops (FFs), and clock cycles, all of which depend on model architecture, quantization, and pruning.
Neural architecture search (NAS) automates model design and can in principle navigate these trade-offs, but most existing methods either optimize for accuracy alone or rely on software-level proxies such as bit operations (BOPs) and floating-point operations (FLOPs) that correlate only loosely with post-synthesis hardware cost.
Closing the gap between the metrics optimized during search and the metrics that dictate deployment requires fast, accurate hardware feedback inside the search loop and a workflow that connects search to physical synthesis end-to-end.

\begin{figure}
    \centering
    \includegraphics[width=0.85\linewidth]{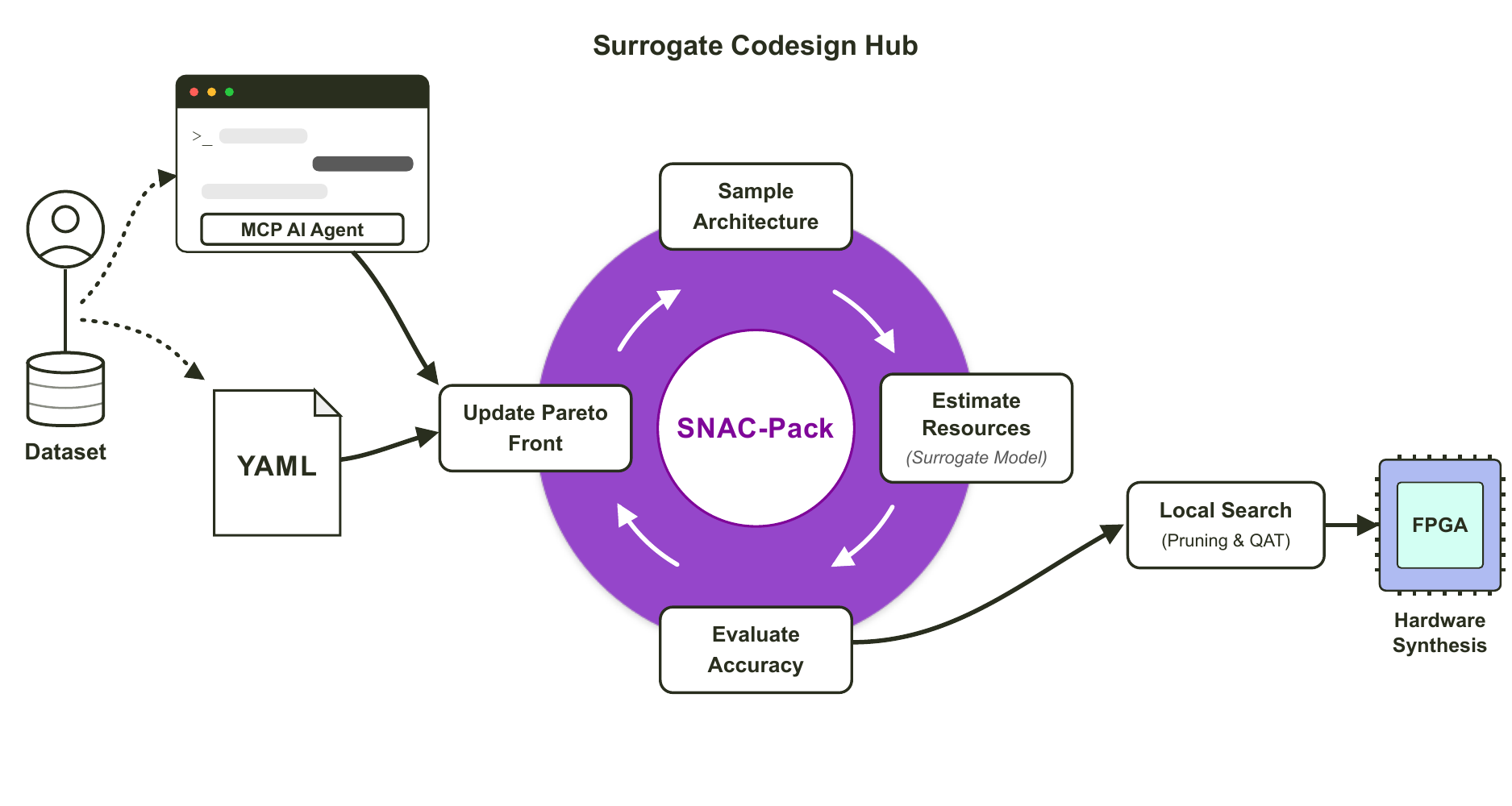}
    \caption{YAML (or optional MCP) drives global multi-objective search with surrogate hardware scores, local QAT and iterative magnitude pruning, and hls4ml synthesis.}
    \label{fig:pipeline_fig}
\end{figure}

We introduce SNAC-Pack 2.0, the next generation of the Surrogate Neural Architecture Codesign framework, an open-source AutoML framework for hardware-aware neural architecture codesign through FPGA deployment (Fig.~\ref{fig:pipeline_fig}; \S\ref{sec:method}). Rather than a new neural architecture search algorithm, SNAC-Pack unifies distributed NAS, surrogate hardware estimation, model compression, and FPGA synthesis into a single end-to-end workflow.
Relative to Neural Architecture Codesign (NAC)~\citep{Weitz_2025}, SNAC-Pack scores global-search trials with learned surrogate estimates of utilization and latency~\citep{Rahimifar_2025} instead of relying on BOPs alone, while preserving multi-objective Optuna~\citep{optuna_2019} search with NSGA-II~\citep{NSGA}, local compression (quantization-aware training and pruning~\citep{quantization_survey,modelcompression}), and hls4ml synthesis~\citep{Schulte:2025mai,hls4ml,fastml_hls4ml}.
Users configure datasets, search spaces, objectives, and hardware in YAML; an optional Model Context Protocol front end exposes the same workflow to agentic clients.

We demonstrate SNAC-Pack on two scientific machine learning tasks with different deployment profiles: jet classification at the Large Hadron Collider~\citep{pierini_2020_3602260}, where sub-microsecond inference is required at the trigger level, and superconducting qubit readout~\citep{diguglielmo2025endtoendworkflowmachinelearningbased}, where readout fidelity must be balanced against strict hardware budgets.
In both cases, SNAC-Pack identifies compact Pareto-optimal architectures that improve the trade-off between task performance and FPGA resource utilization.
SNAC-Pack is open-source and publicly available at \url{https://github.com/fastmachinelearning/nac-opt/tree/automl-2026} ~\citep{dmitri_demler_2026_19202843}.


\section{Related Work}
\label{sec:related}


SNAC-Pack builds on Neural Architecture Codesign (NAC)~\citep{Weitz_2025}: a global search produces a Pareto front, local search refines candidates with quantization-aware training~\citep{quantization_survey} and pruning~\citep{modelcompression}, and hls4ml~\citep{hls4ml,fastml_hls4ml} performs synthesis.
NAC's efficiency objective is BOPs, which only loosely tracks FPGA cost; SNAC-Pack adds rule4ml~\citep{Rahimifar_2025} in the global loop to estimate utilization and latency before invoking full synthesis for every trial.
The following subsections review neural architecture search, hardware-aware NAS, surrogate estimation, and AutoML tooling in that order.

\subsection{Neural Architecture Search}

Neural Architecture Search (NAS) automates architecture design under a search strategy.
\citet{zoph2016neural} introduced reinforcement-learning NAS that was accurate but extremely costly; weight sharing~\citep{pham2018efficient} and DARTS-style relaxation~\citep{liu2018darts} reduced cost but complicate hardware-aware use because supernet rankings can be biased and gradient-based encodings handle non-differentiable FPGA objectives poorly.
Multi-objective evolutionary NAS is surveyed by \citet{9508774}; \citet{lu2019nsga} applied NSGA-II~\citep{NSGA} to trade classification error against complexity and return Pareto fronts, \citet{elsken2018efficient} optimized accuracy and resource cost with network morphism across generations, and \citet{sinha2024multi} extended NSGA-II with zero-cost proxies and diversity preservation to widen Pareto coverage.
Despite this progress, efficiency is still commonly reported with FLOPs, parameter counts, or BOPs, which are cheap but misaligned with reconfigurable-hardware budgets (LUTs, DSPs, BRAM, FFs, clock cycles), motivating hardware-aware NAS.

\subsection{Hardware-Aware NAS}

Hardware-aware NAS targets true deployment cost because FLOP-style proxies correlate poorly with realized performance.
\citet{tan2019mnasnet} incorporated measured mobile latency into the search reward; \citet{wu2019fbnet} avoided repeated on-device calls with additive operator latency tables inside differentiable search; \citet{cai2018proxylessnas} learned latency predictors so candidates could be scored on target hardware without intermediate proxy datasets.
Later work widened the design space: \citet{cai2019once} decoupled supernet training from evolutionary subnet selection using fast accuracy and latency predictors, \citet{wang2020apq} co-searched architecture with pruning and mixed precision, and \citet{benmeziane2023multi} trained rank-preserving surrogates aligned with Pareto fronts rather than absolute error.



Much of this line of work still centers mobile CPUs, GPUs, and edge accelerators, where scalar latency is a workable surrogate.
FPGAs instead bind LUT, DSP, BRAM, and FF utilization together with nontrivial quantization and pruning interactions; \citet{jiang2019accuracy} proposed an early FPGA-aware NAS framework with analytical latency bounds over loop unrolling and pipelining, and \citet{li2021hw} benchmarked FPGA and edge costs on standard NAS spaces, yet both lean on static or analytical models rather than closing the loop for arbitrary architectures, motivating surrogate-based search.

\subsection{Surrogate Models for Hardware Estimation}

Hardware-aware search still needs a per-candidate map to deployment cost, but synthesis or on-device profiling often costs minutes to hours per evaluation, so large search budgets rely on fast learned surrogates instead of repeated ground truth.
For edge CPUs, GPUs, and mobile accelerators, \citet{zhang2021nn} argued fusion-aware kernel-level prediction is needed under graph optimization, \citet{akhauri2024latency} studied transfer of such predictors across devices, and \citet{dudziak2020brp} embedded a GNN over binary architecture relations in NAS to improve sample efficiency when labels are scarce.


Scalar latency surrogates align poorly with FPGAs, where LUT, DSP, FF, and BRAM utilization co-vary with quantization and pruning.
\citet{10.1145/3240765.3264635} trained regressors on HLS code features to predict post-place-and-route performance without invoking HLS in the loop; \citet{10.1145/3489517.3530408} extended this with a GNN over a 40k-program C/C++ benchmark, outperforming HLS-based estimates on resource and timing.
rule4ml~\citep{Rahimifar_2025} specializes the idea to NNs with per-resource regressors on synthesized architectures, forecasting utilization, II, and latency cycles before HLS; wa-hls4ml~\citep{hawks2025wahls4mlbenchmarksurrogatemodels} adds GNN and transformer baselines trained on over 680{,}000 hls4ml-synthesized fully connected and convolutional models as a standardized suite.
SNAC-Pack runs rule4ml with the wa-hls4ml GNN inside the search loop, reserves full hls4ml synthesis for final validation, and keeps the surrogate swappable so future predictors can plug in without changing the search interface.

\subsection{AutoML Frameworks and Tooling}
Most AutoML stacks optimize software-side objectives (for example validation accuracy or throughput) and stop at a trained checkpoint~\citep{feurer2015efficient,jin2023autokeras,zimmer2021auto,wang2021flaml,tang2024autogluon}.
Optuna~\citep{optuna_2019} is a common programmatic backbone with support for distributed workers.
hls4ml translates quantized models to HLS for synthesis but is not a search system.
SNAC-Pack composes Optuna with shared SQLite-backed studies, rule4ml, and hls4ml so FPGA-relevant metrics enter the search loop instead of being evaluated only post hoc.
Related request-to-model agent lines such as AutoMMLab~\citep{yang2025autommlab} motivate our optional Model Context Protocol (MCP) front end, which exposes the same YAML-driven stages as tools for agentic clients; repository documentation lists the full tool surface.

\section{Method}
\label{sec:method}
SNAC-Pack builds on the two-stage global and local search workflow introduced in NAC~\citep{Weitz_2025}: a global stage explores architectural hyperparameters under multi-objective feedback, and a local stage compresses a small set of selected candidates with quantization-aware training (QAT) and pruning before synthesis with hls4ml.
SNAC-Pack extends NAC by implementing hardware surrogate objectives into the global loop (rule4ml~\citep{Rahimifar_2025}; wa-hls4ml predictors~\citep{hawks2025wahls4mlbenchmarksurrogatemodels} are discussed further in \S\ref{sec:related}) so that trial scores reflect FPGA-oriented costs without invoking full synthesis for every candidate.
The framework is designed as an end-to-end workflow for hardware-aware neural architecture search and FPGA deployment, where users configure datasets, search spaces, optimization objectives, hardware estimation settings, and deployment targets through a YAML interface.
Dataset loaders and splits, Optuna study settings, rule4ml/hls4ml-related options, and local-search schedules are part of the same configuration.
The workflow is depicted in Fig.~\ref{fig:pipeline_fig}; concrete hyperparameters for our experiments appear in the case studies so this section stays aligned with the implementation.

\subsection{Global Search}
\label{sec:method-global}
Global search runs multi-objective neural architecture search in Optuna~\citep{optuna_2019} with NSGA-II~\citep{NSGA}.
For each trial, the sampler draws architectural and training hyperparameters (depth, widths, activations, normalization, regularization, and other user-defined knobs) from the search space and builds a Keras model in TensorFlow.
The model is trained for a set number of epochs on the task loss, and a validation metric such as accuracy or readout fidelity is recorded.
When BOPs is an objective, it is computed analytically from the architecture and the configured numeric precision.
If hardware-aware mode is on, the same trained graph is passed to rule4ml's \texttt{MultiModelWrapper} together with an hls4ml-style configuration (board, strategy, precision, reuse factor, and related metadata) so LUT, FF, BRAM, DSP utilization and latency in clock cycles can be estimated quickly instead of invoking Vivado on every candidate.
Each trial returns one scalar per objective in a fixed order, and Optuna persists objectives and study metadata in a relational backend.

NSGA-II evolves a population under Pareto dominance, so the study keeps a set of mutually nondominated trials rather than collapsing everything to a single score.
In our setup, the task metric (accuracy for jet classification; readout fidelity for qubit readout, computed as validation assignment accuracy) is maximized while BOPs, surrogate resource metrics, and surrogate latency are minimized, which matches the deployment goal of keeping task quality high while pushing resource use and cycles down.
Which objectives are turned on is a YAML choice; in the experiments below we use subsets of the task metric, BOPs, mean predicted utilization across LUT/FF/BRAM/DSP, and rule4ml's latency field.

Rule4ml can score Keras or PyTorch models in general~\citep{Rahimifar_2025}; SNAC-Pack passes the trained Keras graph from each trial together with HLS settings to obtain predicted BRAM, DSP, FF, LUT counts or percentages and latency in clock cycles.
For search we aggregate the four utilization percentages into one average-resource objective (their arithmetic mean), while still logging the per-resource columns when we want to inspect constraints or post hoc behavior.
These numbers are inexpensive proxies: they help explore toward promising regions, while the hls4ml synthesis results after local search supply the ground truth we report in the case studies.

Trials may be run back-to-back on one machine or sent out to many workers or Slurm tasks that all point at the same Optuna storage URL.
The bundled workflow uses a shared SQLite study so workers can append completed trials safely; each worker executes the same code path as a single-node run, and ``parallel'' here means independent trials in flight at once, not model-parallel training inside one trial.
Wall time is dominated by short fine-tuning and surrogate calls, not by place-and-route.

\subsection{Local Search and Synthesis}
\label{sec:method-local}

Global search writes a Pareto archive to disk (CSV summaries and YAML checkpoints for selected trials).
SNAC-Pack picks one or more points on that front.
For example, best task performance subject to surrogate budgets is selected and exported as a YAML for the compression stage.

Local search uses SNAC-Pack's combined QAT\,+\,pruning driver (\texttt{tf\_local\_search\_combined}).
For each fixed-point precision pair $(\text{total bits}, \text{integer bits})$ in the local-search configuration, we convert the floating-point seed to a QKeras QAT model, run a QAT warmup for \texttt{qat\_epochs}, then run iterative magnitude pruning with TensorFlow Model Optimization's \texttt{prune\_low\_magnitude}.
Pruning repeats for a configured number of iterations; each pass removes another fraction of the weights that are still nonzero, with the aggressiveness set by the YAML \texttt{pruning\_rate} (concrete values appear in the case-study tables).
Each iteration trains the pruned QAT model for \texttt{epochs\_per\_iteration}, evaluates validation performance, strips pruning wrappers for measurement, logs sparsity and effective BOPs, and applies lottery-ticket-style weight rewinding: pruned weights are zeroed while surviving weights are reset to their post-warmup values before the next iteration.
The best weights seen across iterations for that precision are checkpointed, and the outer loop repeats over every precision pair so we obtain a small family of compressed models rather than a single point estimate.
Optional stratified $k$-fold splits reuse the same schedule per fold and average reported metrics; the fold count is part of the YAML experiment definition.

After local search, chosen checkpoints are exported through hls4ml using the board, I/O style, reuse factor, strategy, and fixed-point formats declared in the experiment configuration.
Vivado reports post-route resource utilization, latency, and initiation interval; those are the numbers we tabulate as ground truth in the case studies and use to check surrogate quality from global search.

\subsection{MCP Interface}
SNAC-Pack optionally exposes the same YAML-driven pipeline through a Model Context Protocol (MCP) layer so agents can plan searches, launch global and local search, and trigger synthesis without hand-editing configuration files.
Tool paths are restricted to the repository root to limit filesystem access; the README in the code repository summarizes the tool catalog, integration modes (native MCP, managed remote MCP, and local OpenAI-style tool calling), and natural-language entry points.










\section{Case Studies}
\label{sec:case-studies}

\subsection{Jet Classification}

To show the effectiveness of SNAC-Pack, we apply it to jet classification, a common and challenging task in high energy physics at the Large Hadron Collider (LHC).
The goal is to accurately classify collision-created jets into one of five categories (light quark, gluon, W boson, Z boson, top quark) based on their kinematic properties.
This is showcased with the hls4ml LHC dataset \cite{pierini_2020_3602260}.
The 8 constituents with the greatest transverse momentum are used per jet.

For this task, we configure SNAC-Pack to search for an optimal multi-layer perceptron (MLP) architecture, with an 8 constituent MLP as a comparative baseline \cite{Odagiu_2024} with the data processed and normalized as done there.
This baseline is chosen, as it is one of the state-of-the-art architectures for this task.
Global search follows \S\ref{sec:method-global} (NSGA-II~\cite{NSGA}) over depth, per-layer widths, activations, batch normalization~\cite{ioffe2015batch}, learning rate, $L_1$ regularization, and dropout (Table~\ref{tab:mlp-parameter-table}, Appendix~\ref{app:jet-search-config}), maximizing accuracy while minimizing mean predicted utilization and surrogate latency.
We use batch size 128, 500 trials, five training epochs per trial, and population size 20, with three-fold cross-validation on one A100 in serial ($\sim$6\,h wall time for global search; $\sim$8\,h for local search across listed bit precisions).

\begin{figure*}[t]
    \centering

    \begin{subfigure}{0.48\textwidth}
        \centering
        \includegraphics[width=\linewidth]{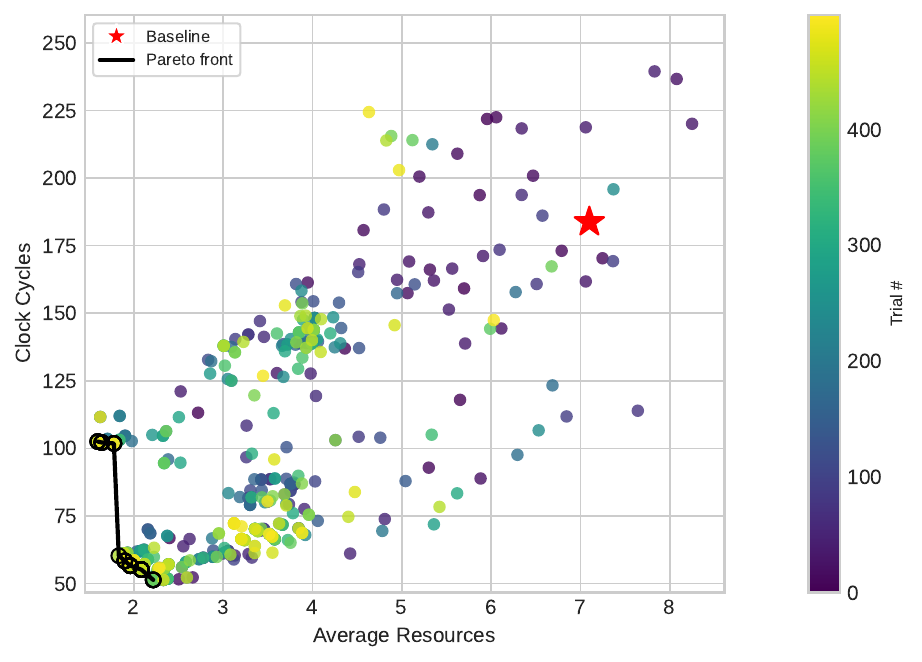}
        \caption{Average resources vs. clock cycles.}
        \label{fig:snac-pack_res_cc}
    \end{subfigure}
    \hfill
    \begin{subfigure}{0.48\textwidth}
        \centering
        \includegraphics[width=\linewidth]{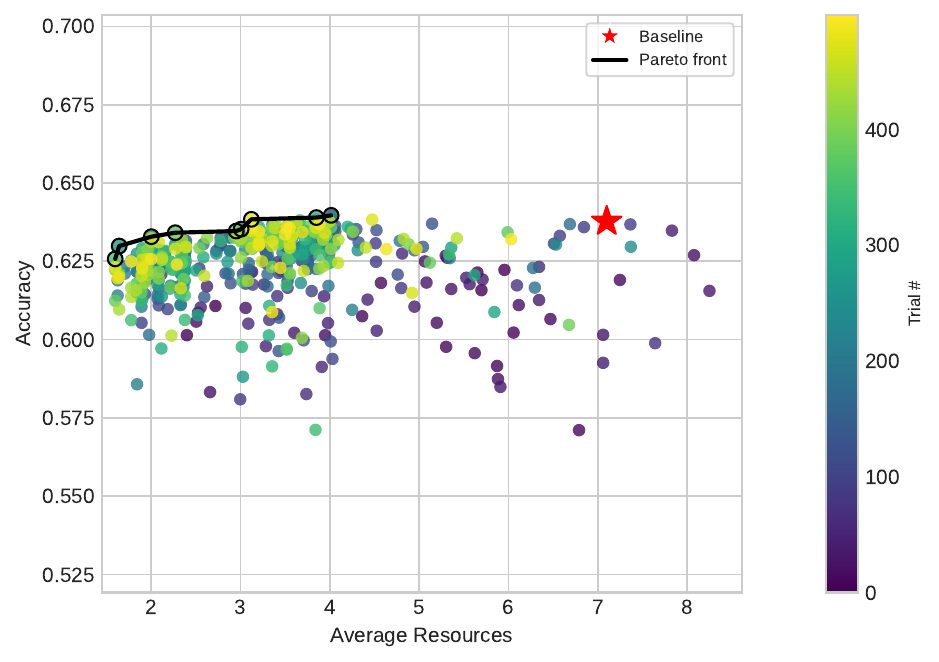}
        \caption{Average resources vs. accuracy.}
        \label{fig:snac-pack_res_acc}
    \end{subfigure}

    \vspace{0.5em}

    \begin{subfigure}{0.48\textwidth}
        \centering
        \includegraphics[width=\linewidth]{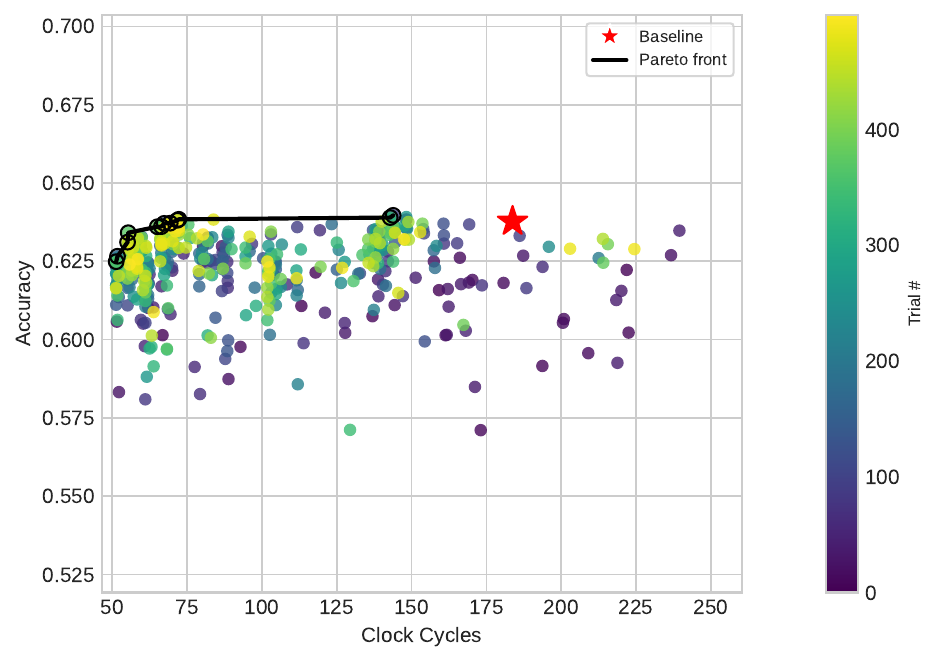}
        \caption{Clock cycles vs. accuracy.}
        \label{fig:snac-pack_cc_acc}
    \end{subfigure}
    \hfill
    \begin{subfigure}{0.48\textwidth}
        \centering
        \includegraphics[width=\linewidth]{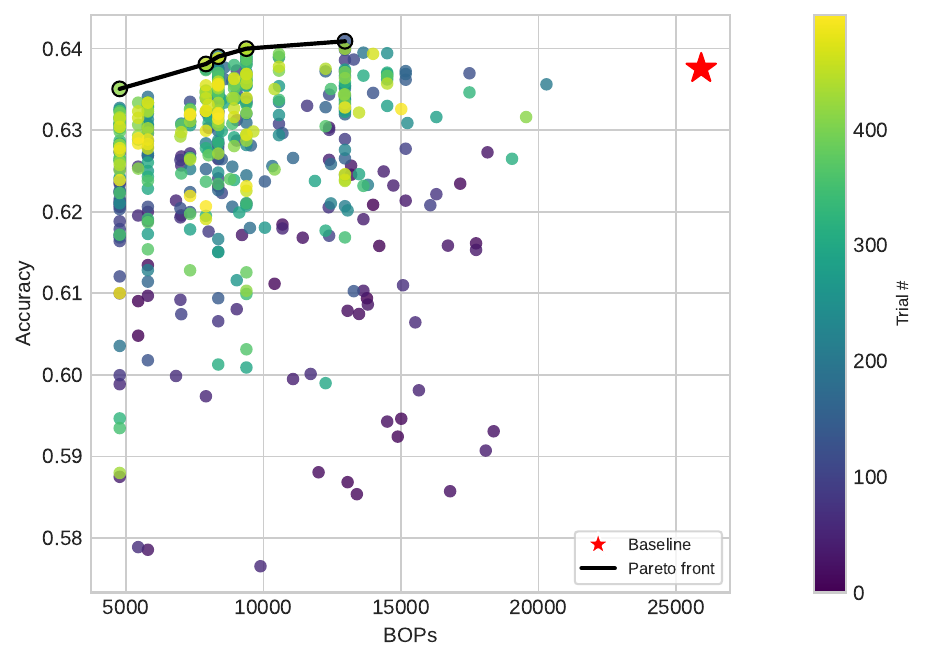}
        \caption{BOPs vs. accuracy (NAC).}
        \label{fig:nac_pareto_front}
    \end{subfigure}

    \caption{
    Pareto fronts obtained during jet classification global search. 
    SNAC-Pack optimizes estimated hardware-aware objectives and accuracy, while NAC optimizes BOPs and accuracy.
    Each point represents a sampled architecture.
    }
    \label{fig:jet_pareto_combined}
\end{figure*}

The results are summarized in Fig.~\ref{fig:jet_pareto_combined}: three SNAC-Pack surrogate Pareto projections (average resources versus clock cycles, average resources versus accuracy, and clock cycles versus accuracy) and, for comparison, an NAC Pareto front in the BOPs-versus-accuracy plane.
For comparison, we also run NAC with the same number of trials and epochs, optimizing solely for BOPs and accuracy.
The resulting Pareto front is shown in Fig.~\ref{fig:nac_pareto_front}.
We take two checkpoints from the SNAC-Pack global Pareto archive to show that the workflow leaves multiple viable deployment choices rather than collapsing to a single scalarized solution.
The Optimal Accuracy SNAC-Pack row is the nondominated trial with highest validation accuracy subject to accuracy greater than 0.638, so it matches or exceeds the baseline~\cite{Odagiu_2024} on that axis.
The Optimal Resource SNAC-Pack row is the Pareto-optimal trial with the lowest BOPs, estimated average resource utilization, and estimated clock cycles among trials with validation accuracy at least 62\%.
Both architectures undergo the same local search and synthesis protocol (Table~\ref{tab:mlp-local-search-table}, Appendix~\ref{app:jet-search-config}) and appear alongside NAC and the baseline in Tables~\ref{tab:mlp-global-comparison} and~\ref{tab:model-comparison-synth}.

Local search follows \S\ref{sec:method-local} with combined QAT and magnitude pruning~\cite{quantization_survey,frankle2019stabilizing,frankle2018lottery}; concrete epochs, pruning rates, and board settings are in Table~\ref{tab:mlp-local-search-table} (Appendix~\ref{app:jet-search-config}).
With hls4ml, the resulting architectures are then synthesized on the Xilinx Virtex UltraScale+ VU13P FPGA, with io\_parallel io\_type, latency strategy, a reuse factor of 1.
Resource utilization and latency are shown in Table~\ref{tab:model-comparison-synth}.

\begin{table}[t]
    \caption{Comparison of model accuracy, BOPs, and estimated hardware metrics from global search.
    The Baseline was optimized for accuracy, NAC for accuracy and BOPs, and SNAC-Pack for accuracy, estimated average resources, and clock cycles.
    Two SNAC-Pack rows reflect different selection rules on the same Pareto archive.
    The best values are reported in bold.}
  \label{tab:mlp-global-comparison}
  \centering
  \footnotesize
  \begin{tabular}{lcccc}
    \toprule
    Model & Accuracy [\%] & BOPs & Est. average resources [\%]& Est. latency (cc) \\
    \midrule
    Baseline \cite{Odagiu_2024}          & 63.77 & 25{,}916 & 7.10 & 183.74 \\
    Optimal NAC \cite{Weitz_2025}      & 63.81 & 7{,}904  & 3.60   & 62.69 \\
    Optimal Accuracy SNAC-Pack & \textbf{63.84} & 8{,}352  & 3.12 & 72.24 \\
    Optimal Resource SNAC-Pack & 62.30 & \textbf{4{,}768} & \textbf{1.96} & \textbf{56.59} \\
    \bottomrule
  \end{tabular}
\end{table}

Table~\ref{tab:mlp-global-comparison} summarizes surrogate scores at export; Table~\ref{tab:model-comparison-synth} gives post-route hardware.
Optimal Resource SNAC-Pack accepts a modest accuracy relaxation but delivers the best latency, II, LUT, FF, and BRAM among the searched rows, whereas Optimal Accuracy SNAC-Pack leads on validation accuracy yet is slower than the baseline after synthesis, illustrating surrogate--Vivado misalignment for latency on that checkpoint.

\begin{table}[t]
  \caption{Hardware resource utilization and latency for the selected models after local search and synthesis.
  The baseline is pruned by 50\% and quantized to 8 bits.
  NAC and both SNAC-Pack checkpoints are synthesized after the same style of local search. CC is the number of clock cycles.
  The SNAC-Pack architectures follow the pruning and QAT schedule in Table~\ref{tab:mlp-local-search-table}.
  The best values are reported in bold.}
  \label{tab:model-comparison-synth}
  \centering
  \scriptsize
  \resizebox{\textwidth}{!}{%
    \begin{tabular}{lcccccc}
      \toprule
      Model & Lat. [ns] (cc) & II [ns] (cc) & DSP & LUT & FF & BRAM \\
      \midrule
      Baseline \cite{Odagiu_2024} & 105 (21) & 5 (1) & 262 (2.1\%) & 155080 (9.0\%) & 25714 (0.7\%) & 4 (0.1\%) \\
      Optimal NAC \cite{Weitz_2025} & 125 (25) & 60 (12) & \textbf{0} & 54075 (3.13\%) & 12016 (0.35\%) & 8 (0.3\%) \\
      Optimal Accuracy SNAC-Pack & 140 (24) & 70 (12) & \textbf{0} & 57728 (3.34\%) & 12605 (0.36\%) & \textbf{0} \\
      Optimal Resource SNAC-Pack& \textbf{50 (10)} & \textbf{5 (1)} & \textbf{0} & \textbf{34472 (1.99\%)} & \textbf{5061 (0.15\%)} & \textbf{0} \\
      \bottomrule
    \end{tabular}%
  }
\end{table}

\subsection{Qubit Readout}

We also evaluate SNAC-Pack on qubit readout classification using superconducting qubit I/Q signal data~\cite{diguglielmo2025endtoendworkflowmachinelearningbased}.
The task is to classify the qubit state from time-series microwave readout signals under strict latency and hardware resource constraints.
The dataset consists of in-phase (I) and quadrature (Q) signals collected from dispersive transmon qubit measurements, where the measured microwave response depends on the underlying qubit state.
Readout fidelity is the validation assignment accuracy (\%): the fraction of readout windows whose predicted binary qubit state matches the label.

For this task, we configure SNAC-Pack to search over compact MLP-based architectures for hls4ml deployment.
Global search follows \S\ref{sec:method-global} with NSGA-II~\cite{NSGA} in Optuna over Table~\ref{tab:qubit-global-parameter-table} (Appendix~\ref{app:qubit-search-config}), maximizing readout fidelity while minimizing rule4ml resource utilization, clock cycles, and BOPs.
We distribute 500 trials (fifty per node) across ten Slurm nodes with one NVIDIA A100 GPU each, five epochs per trial, stratified three-fold cross-validation, and batch size 128 ($\sim$3\,h per node wall time for global search in parallel; $\sim$1.5\,h per node for local search with four precisions across nodes).
This is a large reduction in design space exploration time compared with the manual optimization workflow for this application, where candidate architectures and readout windows were iteratively hand-tuned and synthesized over a period of months~\cite{javi_dse_time}.
We export the highest-fidelity trial as \texttt{best\_model\_for\_local\_search.yaml} for local search under Table~\ref{tab:qubit-local-search-table} (Appendix~\ref{app:qubit-search-config}), then compiled and synthesized with  hls4ml and Vivado as reported in Table~\ref{tab:qubit-model-comparison-synth}.

The baseline model uses a 400 clock cycle readout window starting at clock cycle 100, where the I and Q signal windows are concatenated into an 800-dimensional input.
SNAC-Pack explores smaller readout windows and reduced model capacity to reduce latency and hardware utilization while maintaining readout fidelity.
The search also allows optional removal of hidden activations to evaluate whether nonlinearities improve performance enough to justify additional deployment cost.

\begin{figure*}[t]
    \centering

    \begin{subfigure}{0.48\textwidth}
        \centering
        \includegraphics[width=\linewidth]{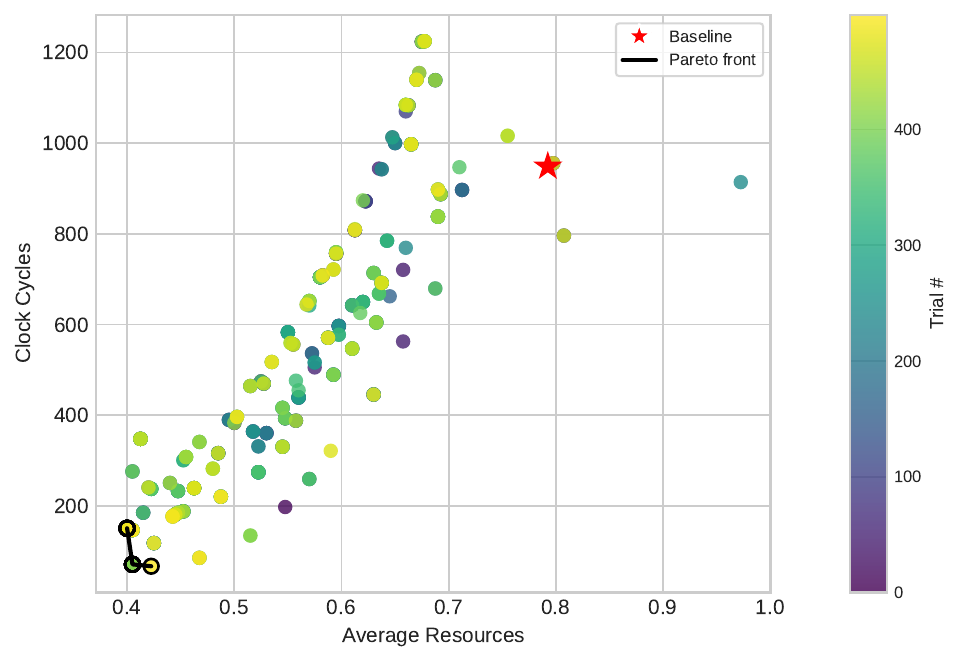}
        \caption{Average resources vs. clock cycles.}
        \label{fig:snac-pack_qubit_res_cc}
    \end{subfigure}
    \hfill
    \begin{subfigure}{0.48\textwidth}
        \centering
        \includegraphics[width=\linewidth]{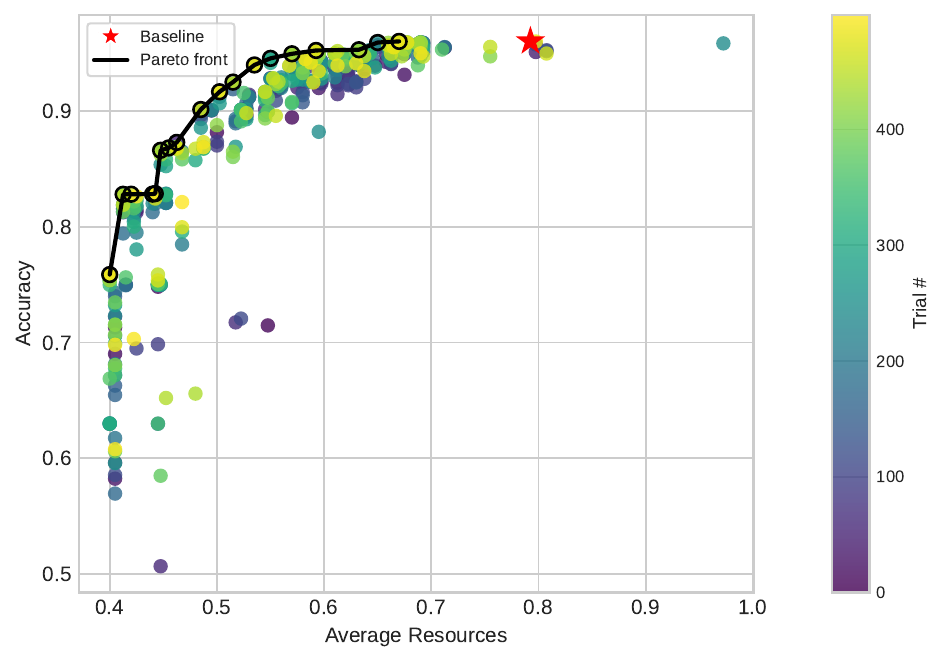}
        \caption{Average resources vs. accuracy.}
        \label{fig:snac-pack_qubit_res_acc}
    \end{subfigure}

    \vspace{0.5em}

    \begin{subfigure}{0.48\textwidth}
        \centering
        \includegraphics[width=\linewidth]{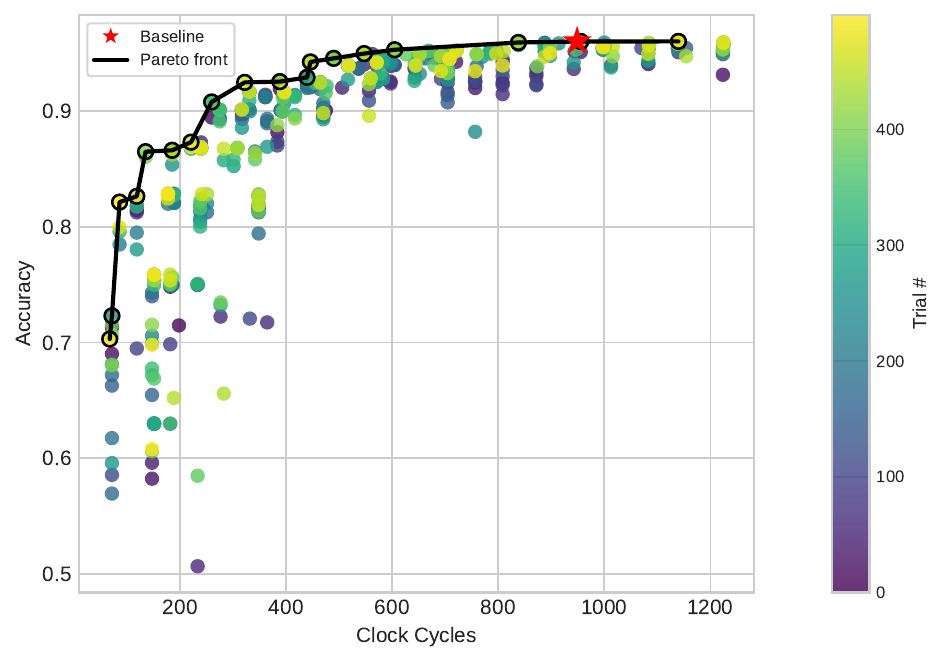}
        \caption{Clock cycles vs. accuracy.}
        \label{fig:snac-pack_qubit_cc_acc}
    \end{subfigure}
    \hfill
    \begin{subfigure}{0.48\textwidth}
        \centering
        \includegraphics[width=\linewidth]{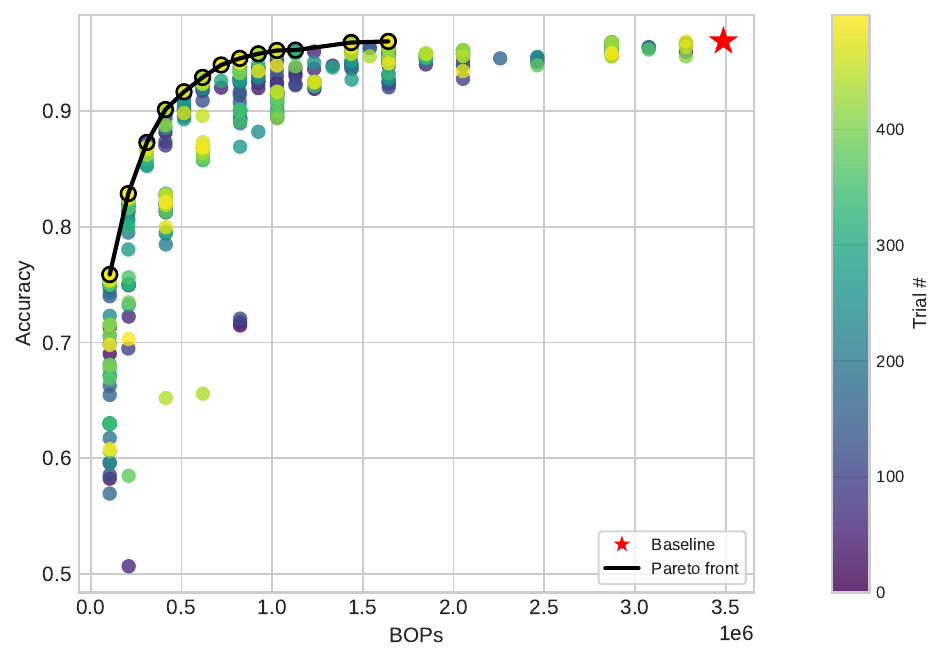}
        \caption{BOPs vs. accuracy.}
        \label{fig:snac-pack_qubit_bops_acc}
    \end{subfigure}

    \caption{
    Pareto fronts obtained during qubit readout global search. 
    SNAC-Pack optimizes estimated hardware-aware objectives, readout fidelity, and BOPs.
    Each point represents a sampled architecture.
    }
    \label{fig:qubit_pareto_combined}
\end{figure*}

Figure~\ref{fig:qubit_pareto_combined} summarizes surrogate trade-offs from the completed study: each panel projects the sampled architectures in the plane of validation fidelity against one hardware-aware axis (average resources, clock cycles, or BOPs), so the cloud of points traces how much fidelity must be relaxed as estimated utilization, latency, or operator count is pushed down.
Unlike the jet experiment, we report a single SNAC-Pack checkpoint that matches the export rule above. The trial with highest validation fidelity is taken from the archive and carried through the same local search and synthesis protocol, which appears alongside the baseline in Tables~\ref{tab:qubit-global-comparison} and~\ref{tab:qubit-model-comparison-synth}.

\begin{table}[t]
  \caption{Comparison of readout fidelity, BOPs, and estimated hardware metrics from global search.
  Readout fidelity denotes validation assignment accuracy (\%) on the scored split.
  The baseline was tuned for fidelity alone; SNAC-Pack was searched for fidelity together with estimated average resources and clock cycles.
  The best values are reported in bold.}
  \label{tab:qubit-global-comparison}
  \centering
  \footnotesize
  \begin{tabular}{lcccc}
    \toprule
    Model & Fidelity [\%] & BOPs & Est. average resources [\%] & Est. latency (cc) \\
    \midrule
    Baseline \cite{diguglielmo2025endtoendworkflowmachinelearningbased} & \textbf{96.06} & 3{,}487{,}250 & 0.79 & 948.67 \\
    Optimal SNAC-Pack & 95.21 & \textbf{1{,}026{,}048} & \textbf{0.61} & \textbf{810.36} \\
    \bottomrule
  \end{tabular}
\end{table}

\begin{table}[t]
  \caption{Hardware resource utilization and latency estimates for qubit readout models synthesized with hls4ml.
  The baseline model uses a 400 clock cycle readout window with an 800-dimensional input.
  The SNAC-Pack model is synthesized after local search with iterative pruning and quantization-aware training.
  CC denotes the number of clock cycles.
  The best values are reported in bold.}
  \label{tab:qubit-model-comparison-synth}
  \centering
  \scriptsize
  \resizebox{\textwidth}{!}{%
    \begin{tabular}{lcccccc}
      \toprule
      Model & Lat. [ns] (cc) & II [ns] (cc) & DSP & LUT & FF & BRAM \\
      \midrule
      Baseline & 19.35 (6) & 16.1 (5) & \textbf{0} & 15045 (5.49\%) & 3656 (0.67\%) & \textbf{0} \\
      Optimal SNAC-Pack & \textbf{12.9 (4)} & \textbf{12.9 (4)} & \textbf{0} & \textbf{6996 (2.55\%)} & \textbf{2330 (0.43\%)} & \textbf{0} \\
      \bottomrule
    \end{tabular}%
  }
\end{table}

Table~\ref{tab:qubit-global-comparison} makes the surrogate trade-off explicit for that exported design: relative to the baseline tuned for fidelity alone, readout fidelity moves from $96.06\%$ to $95.21\%$ while BOPs fall from $3{,}487{,}250$ to $1{,}026{,}048$, estimated average resource utilization drops from $0.79\%$ to $0.61\%$, and the rule4ml clock-cycle latency estimate decreases from $948.67$ to $810.36$.
The task-metric loss is modest compared with the reduction in operator count and the surrogate hardware scores, consistent with the broad achievable region illustrated in Fig.~\ref{fig:qubit_pareto_combined}.

After local search and Vivado synthesis, Table~\ref{tab:qubit-model-comparison-synth} shows that the same checkpoint also occupies a cheaper post-route point than the baseline: latency and initiation interval shrink to $12.9\,\mathrm{ns}$ (four clock cycles) from $19.35\,\mathrm{ns}$ (six cycles), while LUT and FF counts are reduced to roughly half of the baseline footprint (DSP and BRAM remain at zero in both rows).
Because the search varies readout windowing and layer sizes, these results additionally indicate that competitive fidelity can be maintained away from the fixed $400$-cycle baseline window while still improving on-chip timing and area.

\section{Conclusion}
This work introduced SNAC-Pack, an open-source AutoML pipeline for hardware-aware neural architecture search, model compression, and FPGA deployment.
Built on Optuna and parallel trial workers, SNAC-Pack extends NAC by scoring global search trials with rule4ml surrogate estimates of utilization and latency.

On LHC jet classification, NSGA-II produces a Pareto set of models instead of a single best design. The resource-focused SNAC-Pack model does better than the baseline and NAC on post-route LUT, flip-flop, BRAM, latency, and initiation interval after the same compression setup (Table~\ref{tab:model-comparison-synth}). The accuracy-focused model stays strong on validation accuracy but is slower than the baseline after synthesis, which shows that surrogate rank during search can disagree with Vivado timing and that predictors can be improved.
On superconducting qubit readout, we export the trial with highest validation fidelity and run the same QAT and pruning pipeline. Readout fidelity drops slightly compared to the baseline while BOPs and surrogate hardware scores improve, and post-route latency and area on the ZCU102 beat the tabulated baseline (Tables~\ref{tab:qubit-global-comparison} and~\ref{tab:qubit-model-comparison-synth}). By using SNAC-Pack, the time required to find an optimal model was reduced from months originally spent hand-tuning models~\citep{javi_dse_time} to a few hours of automated search. 
Users configure datasets, objectives, and hardware settings through YAML and an optional MCP layer allows for the use of the same sandboxed tools.
Future work includes improving and calibrating surrogate estimation, evaluating search variance across repeated optimization runs, expanding additional support to PyTorch, and expanding search spaces as predictors support them.

\clearpage

\section*{Acknowledgements}
Fermilab Report Number FERMILAB-CONF-26-0339-CSAID.

This manuscript has been authored by Fermi Forward Discovery Group, LLC under Contract No. 89243024CSC000002 with the U.S. Department of Energy, Office of Science, Office of High Energy Physics.

BH and NT are also supported under the DOE Early Career Research program under Award No. DE-0000247070.

BH, JD, and NT are supported by the U.S. Department of Energy (DOE), Office of Science, Office of Advanced Scientific Computing Research under the ``Real-time Data Reduction Codesign at the Extreme Edge for Science'' Project(DE-FOA-0002501).

JD is also supported by the DOE, Office of Science, Office of High Energy Physics Early Career Research program under Grant No. DE-SC0021187, and the U.S. National Science Foundation (NSF) Harnessing the Data Revolution (HDR) Institute for Accelerating AI Algorithms for Data Driven Discovery (A3D3) under Cooperative Agreement No. PHY-2117997.

JW is supported by a WATCHEP fellowship sponsored by the DOE, Office of High-Energy Physics under Award No. DE-SC-0023527.

BH and NT are supported by DOE BIA Microelectronics Science Research Center and the Transformational Model Consortium.

This research used resources of the National Energy Research Scientific Computing Center (NERSC), a Department of Energy User Facility (project amsc011-2026).

\newpage
\bibliography{references}

\newpage

\newpage
\appendix
\section{Supplemental Section}

\subsection{Jet classification search space and configuration}
\label{app:jet-search-config}

\begin{table}[H]
  \caption{Comprehensive parameter space for jet classification.}
  \label{tab:mlp-parameter-table}
  \centering
  \small
  \renewcommand{\arraystretch}{1.2}
  \begin{tabular}{ll}
    \toprule
    \textbf{Parameter} & \textbf{Space} \\
    \midrule
    Number of layers        & \{4, 5, 6, 7, 8\} \\
    Hidden units per layer  & \\
    \quad Layer 1 & \{64, 120, 128\} \\
    \quad Layer 2 & \{32, 60, 64\} \\
    \quad Layer 3 & \{16, 32\} \\
    \quad Layer 4 & \{32, 64\} \\
    \quad Layer 5 & \{32, 64\} \\
    \quad Layer 6 & \{32, 64\} \\
    \quad Layer 7 & \{16, 32\} \\
    \quad Layer 8 & \{32, 44, 64\} \\
    Activation function     & \{ReLU, Tanh, Sigmoid\} \\
    Batch normalization     & \{True, False\} \\
    Learning rate           & \{0.0010, 0.0015, 0.0020\} \\
    L1 regularization       & \{0.0, $10^{-6}$, $10^{-5}$, $10^{-4}$\} \\
    Dropout rate            & \{0.0, 0.05, 0.1\} \\
    \bottomrule
  \end{tabular}
\end{table}

\begin{table}[H]
  \caption{Local search and synthesis configuration for jet classification.}
  \label{tab:mlp-local-search-table}
  \centering
  \small
  \renewcommand{\arraystretch}{1.2}
  \begin{tabular}{ll}
    \toprule
    \textbf{Parameter} & \textbf{Value / Space} \\
    \midrule
    Warm-up epochs & 5 \\
    Pruning iterations & 10 \\
    Pruning epochs per iteration & 10 \\
    Pruning rate per iteration & 20\% \\
    Quantization precision & \{(32,16), (16,6), (8,3), (4,1)\} \\
    FPGA board & VU13P \\
    HLS io\_type & io\_parallel \\
    HLS strategy & Latency \\
    Reuse factor & 1 \\
    \bottomrule
  \end{tabular}
\end{table}

\subsection{Qubit readout search space and configuration}
\label{app:qubit-search-config}

\begin{table}[H]
  \caption{Global search parameter space for qubit readout classification.}
  \label{tab:qubit-global-parameter-table}
  \centering
  \small
  \renewcommand{\arraystretch}{1.2}
  \begin{tabular}{ll}
    \toprule
    \textbf{Parameter} & \textbf{Space} \\
    \midrule
    Hidden units & \{2, 4\} \\
    Hidden activation & \{ReLU, LeakyReLU, None\} \\
    Normalization & \{BatchNorm, None\} \\
    Block type & \{None\} \\
    Number of blocks & \{1\} \\
    MLP head layers & \{2\} \\
    Output dimension & \{1\} (binary readout; scalar labels) \\
    Output activation & \{None (logits; binary cross-entropy)\} \\
    Readout window size &
    \{25, 50, 75, ..., 400\} \\
    Window start location &
    \{0, 25, 50, ..., 725\} \\
    \bottomrule
  \end{tabular}
\end{table}

\begin{table}[H]
  \caption{Local search and synthesis configuration for qubit readout classification.}
  \label{tab:qubit-local-search-table}
  \centering
  \small
  \renewcommand{\arraystretch}{1.2}
  \begin{tabular}{ll}
    \toprule
    \textbf{Parameter} & \textbf{Value / Space} \\
    \midrule
    Warm-up epochs & 20 \\
    Pruning iterations & 10 \\
    Pruning epochs per iteration & 20 \\
    Pruning rate per iteration & 20\% \\
    Quantization precision & \{(32,16), (16,6), (8,3), (4,1)\} \\
    FPGA board & ZCU102 \\
    HLS strategy & Latency \\
    Reuse factor & 8 \\
    \bottomrule
  \end{tabular}
\end{table}

\end{document}